\def\year{2020}\relax
\documentclass[letterpaper]{article} 
\usepackage{aaai20}  
\usepackage{times}  
\usepackage{helvet} 
\usepackage{courier}  
\usepackage[hyphens]{url}  
\usepackage{graphicx} 
\usepackage{graphicx}  
\usepackage{amsmath}
\usepackage{booktabs}
\usepackage{algorithm}
\usepackage{algpseudocode}
\usepackage{amsthm}
\usepackage{amssymb}
\usepackage{tikz}
\usepackage{verbatim}
\usepackage{subfig}
\usepackage{booktabs}
\usepackage{multirow}
\usepackage{siunitx}

\DeclareMathOperator*{\argmax}{arg\,max}
\usepackage{lmodern,caption}
\usepackage{array,booktabs,makecell}

\newtheoremstyle{own}%
    {3pt}
    {3pt}
    {}
    {}
    {\color{black}\bfseries}
    {:}

    {}

\theoremstyle{definition}
\newtheorem{example}{Example}

\theoremstyle{definition}
\newtheorem{definition}{Definition}
\def\mathcalc{}

\title{Explainable Reinforcement Learning Through a Causal Lens}
\author{\Large 
\textbf{Prashan Madumal, Tim Miller, Liz Sonenberg, Frank Vetere}\\ 
The University of Melbourne\\ 
Victoria, Australia\\
pmathugama@student.unimelb.edu.au,
\{tmiller, l.sonenberg, f.vetere\}@unimelb.edu.au 
}
 
\begin{document}

\maketitle

\begin{abstract}
Prominent theories in cognitive science propose that humans understand and represent the knowledge of the world through causal relationships. In making sense of the world, we build \emph{causal models} in our mind to encode cause-effect relations of events and use these to \emph{explain} why new events happen by referring to counterfactuals --- things that did not happen. In this paper, we use causal models to derive causal explanations of the behaviour of model-free reinforcement learning agents. We present an approach that learns a \emph{structural causal model} during reinforcement learning and encodes causal relationships between variables of interest. This model is then used to generate explanations of behaviour based on counterfactual analysis of the causal model. We computationally evaluate the model in 6 domains and measure performance and task prediction accuracy. We report on a study with \textbf{120} participants who observe agents playing a real-time strategy game (Starcraft II) and then receive explanations of the agents' behaviour. We investigate: 1) participants' understanding gained by explanations through task prediction; 2) explanation satisfaction and 3) trust. Our results show that causal model explanations perform better on these measures compared to two other baseline explanation models. 
\end{abstract}

Driven by lack of trust from users and proposed regulations, there are many calls for Artificial Intelligence (AI) systems to become more transparent, interpretable and explainable. This has renewed the interest in Explainable AI (XAI), which has been explored since the expert systems era~\cite{chandrasekaran1989explaining}. A key pillar of XAI is~\emph{explanation}, a justification given for decisions and actions of the system.

However, much research and practice in XAI pays little attention to \emph{people} as intended users of these systems~\cite{miller2018explanation}. If we are to build systems that are capable of providing `good' explanations, it is plausible that explanation models should mimic models of human explanation~\cite{deGraaf}. Thus, to build XAI models it is essential to begin with a strong understanding of how people define, generate, select and evaluate explanations. 


There is a wealth of pertinent literature in cognitive psychology that explore the nature of explanations and how people understand them. As humans, we view the world through a causal lens~\cite{sloman2005causal}, building mental models with causal relationships to act in the world, to understand new events and also to \emph{explain} events. Importantly, causal models give people the ability to consider \emph{counterfactuals} --- events that did not happen, but could have under different situations. Although this notion of causal explanation is also backed by literature in philosophy and social psychology~\cite{hilton2007causal}, causality and counterfactuals are only just becoming more prevalent in XAI. Further, compared to the burst of XAI research in supervised learning, explainability in model-free reinforcement learning is hardly explored.

We introduce an \emph{action influence} model for model-free reinforcement learning (RL) agents and provide a formalisation of the model using structural causal models~\cite{halpern2005causes}. Action influence models approximate the causal model of the environment relative to actions taken by an agent. Our approach differs from previous work in explainable RL in that we use causal models to generate \emph{contrastive} explanations for \emph{why} and \emph{why not} questions, which previous models lack. Given assumptions about the direction of causal relationships between variables, during the policy learning process, we also learn the quantitative influences that actions have on variables. Which enable our model to reason approximately about counterfactual states and actions. We define how to generate explanations for `why?' and `why not?' questions from the action influence model. We define \emph{minimally complete} explanations taking inspiration from social psychology literature~\cite{mcclure1997you}.

We computationally evaluated our approach on 6 RL benchmarks domains using 6 different RL algorithms. Results indicate that these models are robust and accurate enough to perform task prediction~\cite[p.12]{hoffman2018metrics} with a negligible performance impact. We conducted a human study using the implemented model for RL agents trained to play the real-time strategy game \emph{Starcraft II}. Experiments were run for \textbf{120} participants, in which we evaluated the participants' performance in task prediction, explanation satisfaction, and trust. Results show that our  model performs better than the tested baseline, but its impact on trust is not statistically significant.

The main contribution of this paper is twofold: 1) We introduce and formalise the \emph{action influence} model based on structural causal models and present definitions to generate explanations; 2) We conduct a between-subject human study to evaluate the proposed model with baselines.

\section{Related Work}


There exists a substantial body of literature that explores explaining the policies and actions of Markov Decision Processes (MDP), though most of them do not explicitly focus on reinforcement learning.~\citeauthor{elizalde2008policy}~\shortcite{elizalde2009expert} generated explanations by selecting and using `relevant' variables of states of factored MDPs, evaluated by domain experts. Taking the long term effect an action has, ~\citeauthor{khan2009minimal}~\shortcite{khan2009minimal} proposed generating sufficient and minimal explanations for MDPs using domain independent templates.

Policy explanations in human-agent interaction settings have been used to achieve transparency~\cite{hayes2017improving} and provide summaries of the policies~\cite{amir2018highlights}. Explanation in reinforcement learning has been explored, using interactive RL to generate explanations from instructions of a human~\cite{fukuchi2017autonomous} and to provide contrastive explanations~\cite{van2018contrastive}. Soft decision trees have been used to generate more interpretable policies~\cite{YouriC}, and reward decomposition has been utilized to provide minimum sufficient explanations in RL~\cite{ZoeJuozapaitis}. However, these explanations are not based on an underlying causal model.



Other work on causal explanation has focused on scientific explanations~\cite{salmon1984scientific} and explanations using causal trees~\cite{nielsen2012explanation}. Although some recent work has emphasized the importance of causal explanation for explainable AI systems~\cite{miller2018explanation,miller2018contrastive,madumal2019grounded,madumal2019explainable}, work on generating explanations from causal explanation models for MDPs and RL agents have been absent. 


\section{Causal Models for Explanations}
\label{sec:causal}
In this section, we introduce the \emph{action influence model}, which is based on  \emph{structural causal models} of \citeauthor{halpern2005causes}~\shortcite{halpern2005causes}. We first introduce the Starcraft II domain, a partially observable real-time strategy game environment with a large state and action space. For the purpose of implementing RL agents for explanation, we use a toned-down version of the full Starcraft II 1v1 match (an adversarial scenario) with 4 actions and 9 state variables for the agent's model (see Figure \ref{fig1}). In the following sections we use this Starcraft II scenario accompanied by Figure \ref{fig1} as our running example.

\subsection{Preliminaries : Structural Causal Models}

Structural causal models (SCMs) \cite{halpern2005causes} represent the world using random variables, divided into exogenous (external) and endogenous (internal), some of which might have causal relationships which each other. These relationships can be described with a set of \emph{structural equations}. Formally, a \emph{signature} $\mathcalc{S}$ is a tuple $\left ( \mathcal{U}, \mathcal{V}, \mathcal{R} \right )$, where $\mathcal{U}$ is the set of exogenous variables, $\mathcal{V}$ the set of endogenous variables, and $\mathcal{R}$ is a function that denotes the range of values for every variable $\mathcal{Y} \in   \mathcal{U} \cup \mathcal{V} $.

\begin{definition}\label{def1}
A \emph{structural causal model} is a tuple $M = \left (\mathcal{S}, \mathcal{F}  \right )$, where $\mathcal{F}$ denotes a set of structural equations, one for each $X \in \mathcal{V}$, such that $F_X : \left ( \times_{U \in \mathcal{U}}\mathcal{R}(U) \right )\times\left ( \times_{Y \in \mathcal{V} -\left \{ X \right \}}\mathcal{R}(Y) \right )\rightarrow \mathcal{R}(X)$ give the value of $X$ based on other variables in $\mathcal{U} \cup \mathcal{V}$. That is, the equation $F_X$ defines the value of $X$ based on some other variables in the model.
\end{definition}

A \emph{context} $\vec{u}$ is a vector of unique values of each exogenous variable $u \in \mathcal{U}$. A \emph{situation} is defined as a model/context pair $\left (M, \vec{u}  \right )$. An \emph{instantiation} is defined by assigning variables the values corresponding to those defined by their structural equations. An \emph{actual cause} of an event $\varphi$ is a vector of endogenous variables and their values such that there is some counterfactual context in which the variables in the cause are different and the event $\varphi$ does not occur. An explanation is those causes that an explainee does not already know.
For a more complete review of SCM's we direct the reader to~\cite{halpern2005causes}.
%



\subsection{Causal Models for Reinforcement Learning Agents}


Our intent in this paper is not to provide explanations of \emph{evidence} from the environment, but to provide explanations of the agent's behaviour based on the knowledge of how actions influence the environment. As such, we extend the notion of SCMs to include actions as part of the causal relationships.

We incorporate  \emph{action influence} models for MDP-based RL agents, extending SCMs with the addition of actions. An MDP is a tuple $\left ( \mathcal{S}, \mathcal{A}, \mathcal{T}, \mathcal{R} \gamma   \right )$, where $S$ and $A$ give state and action spaces respectively (here we assume the state and action space is finite and state features are described by a set of variables $\phi$); $\mathcal{T} = \left \{ P_{sa} \right \}$ a set of state transition functions ($P_{sa}$ denotes state transition distribution of taking action $a$ in state $s$); $\mathcal{R} : \mathcal{S} \times \mathcal{A} \rightarrow \mathbb{R}$ is a reward function and $\gamma = \left [0, 1  \right )$ is a discount factor. The objective of an RL agent is to find a policy $\pi$ that maps states to actions maximizing the expected discounted sum of rewards. We define the action influence model for RL agents as follows.

Formally, a signature $\mathcalc{S_{a}}$ for an action influence model is a tuple $(\mathcal{U}, \mathcal{V}, \mathcal{R}, \mathcal{A})$, in which $\mathcal{U}$, $\mathcal{V}$, and $\mathcal{R}$ are as in SCMs, and $\mathcal{A}$ is the set of actions.

\begin{definition}\label{def2}
An \emph{action influence model} is a tuple $\left (  \mathcalc{S_{a}}, \mathcal{F}\right)$, where $\mathcalc{S_{a}}$ is as above, and $\mathcal{F}$ is the set of structural equations, in which we have multiple for each $X \in \mathcal{V}$ --- one for each \emph{unique} action set that influences $X$. A function $F_{X.A}$, for $A \in \mathcal{A}$, defines the causal effect on $X$ from applying action $A$. The set of \emph{reward variables} $\mathcalc{X_r} \subseteq \mathcal{V}$ are defined by the set of nodes with an out-degree of 0; that is, the set of sink nodes.
\end{definition}


We define the \emph{actual instantiation} of a model $M$ as $M_{\vec{\mathcal{V}}\leftarrow \vec{S}}$, in which $\vec{S}$ is the vector of state variable values from an MDP. In an actual instantiation, we set the values of all state variables in the model, effectively making the exogenous variables irrelevant. 


Figure~\ref{fig1} shows the graphical representation of Definition \ref{def2} as an action influence graph of the Starcraft II agent described in the previous section, with exogenous variables hidden. These \emph{action influence models} are SCMs except that each edge is associated with an action. In the action influence model, each state variable has a \emph{set} of structural equations: one for each \emph{unique} incoming action. As an example, from Figure \ref{fig1}, variable $A_n$ is causally influenced by $S$ and $B$ only when action $A_m$ is executed, thus the structural equation $\mathcalc{F_{A_n.A_m}}\left ( S, B \right )$ captures that relationship.

\section{Explanation Generation}
In this section, we present definitions that generate explanations from an action influence model. 
The process of explanation generation has 3 phases: 1) defining the qualitative causal relationships of variables as an action influence model; 2) learning the structural equations during RL; and 3) generating \emph{explanans} from SCMs using the definitions given below.

We define an \emph{explanation} as a pair that consist of: 1) an \emph{explanandum}, the event to be explained; and 2) an \emph{explanan}, the subset of causes given as the explanation~\cite{miller2018explanation}. Consider the example `Why did you do \emph{P}?' and the explanation `Because of \emph{Q}'. Here, the \emph{explanandum} is \emph{P} and \emph{explanan} is \emph{Q}. Identifying the \emph{explanandum} from a question is not a trivial task. In this paper, we define explanations for questions of the form `Why $A$?' or `Why not $A$?', where $A$ is an action. In the context of a RL agent we define a \emph{complete explanan} below.

\begin{definition}\label{def5}
A \emph{complete explanan} for an action $a$ under the actual instantiation $M_{\vec{\mathcal{V}}\leftarrow \vec{S}}$ is a tuple $\left (  \vec{X_r}=\vec{x_r}, \vec{X_h}=\vec{x_h}, \vec{X_i}=\vec{x_i}\right )$, in which $ \vec{X_r}$ is the vector of reward variables reached by following the causal chain of the graph to sink nodes; $\vec{X_h}$ the vector of variables of the head node of action $a$, $\vec{X_i}$ the vector of intermediate nodes between head and reward nodes, and $\vec{x_r}$, $\vec{x_h}$, $\vec{x_i}$ gives the values of these variables under $M_{\vec{\mathcal{V}}\leftarrow \vec{S}}$. 
\end{definition}

\begin{figure}[tb]

\centering
\resizebox {\columnwidth} {!} {

\tikzset{every picture/.style={line width=0.75pt}} 

\begin{tikzpicture}[x=0.75pt,y=0.75pt,yscale=-1,xscale=1]

\draw  [color={rgb, 255:red, 0; green, 0; blue, 0 }  ,draw opacity=1 ] (3.8,120.6) .. controls (3.8,106.79) and (14.99,95.6) .. (28.8,95.6) .. controls (42.61,95.6) and (53.8,106.79) .. (53.8,120.6) .. controls (53.8,134.41) and (42.61,145.6) .. (28.8,145.6) .. controls (14.99,145.6) and (3.8,134.41) .. (3.8,120.6) -- cycle ;
\draw  [color={rgb, 255:red, 0; green, 0; blue, 0 }  ,draw opacity=1 ][fill={rgb, 255:red, 104; green, 104; blue, 104 }  ,fill opacity=0.11 ] (96.53,170) .. controls (96.53,156.19) and (107.73,145) .. (121.53,145) .. controls (135.34,145) and (146.53,156.19) .. (146.53,170) .. controls (146.53,183.81) and (135.34,195) .. (121.53,195) .. controls (107.73,195) and (96.53,183.81) .. (96.53,170) -- cycle ;
\draw  [color={rgb, 255:red, 0; green, 0; blue, 0 }  ,draw opacity=1 ][fill={rgb, 255:red, 155; green, 155; blue, 155 }  ,fill opacity=1 ] (198,110.67) .. controls (198,96.86) and (209.19,85.67) .. (223,85.67) .. controls (236.81,85.67) and (248,96.86) .. (248,110.67) .. controls (248,124.47) and (236.81,135.67) .. (223,135.67) .. controls (209.19,135.67) and (198,124.47) .. (198,110.67) -- cycle ;
\draw  [color={rgb, 255:red, 0; green, 0; blue, 0 }  ,draw opacity=1 ][fill={rgb, 255:red, 155; green, 155; blue, 155 }  ,fill opacity=1 ] (97.73,69.6) .. controls (97.73,55.79) and (108.93,44.6) .. (122.73,44.6) .. controls (136.54,44.6) and (147.73,55.79) .. (147.73,69.6) .. controls (147.73,83.41) and (136.54,94.6) .. (122.73,94.6) .. controls (108.93,94.6) and (97.73,83.41) .. (97.73,69.6) -- cycle ;
\draw  [color={rgb, 255:red, 0; green, 0; blue, 0 }  ,draw opacity=1 ] (197,187.67) .. controls (197,173.86) and (208.19,162.67) .. (222,162.67) .. controls (235.81,162.67) and (247,173.86) .. (247,187.67) .. controls (247,201.47) and (235.81,212.67) .. (222,212.67) .. controls (208.19,212.67) and (197,201.47) .. (197,187.67) -- cycle ;
\draw  [color={rgb, 255:red, 0; green, 0; blue, 0 }  ,draw opacity=1 ] (197,28.67) .. controls (197,14.86) and (208.19,3.67) .. (222,3.67) .. controls (235.81,3.67) and (247,14.86) .. (247,28.67) .. controls (247,42.47) and (235.81,53.67) .. (222,53.67) .. controls (208.19,53.67) and (197,42.47) .. (197,28.67) -- cycle ;
\draw  [color={rgb, 255:red, 0; green, 0; blue, 0 }  ,draw opacity=1 ] (197,270.67) .. controls (197,256.86) and (208.19,245.67) .. (222,245.67) .. controls (235.81,245.67) and (247,256.86) .. (247,270.67) .. controls (247,284.47) and (235.81,295.67) .. (222,295.67) .. controls (208.19,295.67) and (197,284.47) .. (197,270.67) -- cycle ;
\draw  [color={rgb, 255:red, 0; green, 0; blue, 0 }  ,draw opacity=1 ][fill={rgb, 255:red, 155; green, 155; blue, 155 }  ,fill opacity=1 ] (324.33,198.33) .. controls (324.33,184.53) and (335.53,173.33) .. (349.33,173.33) .. controls (363.14,173.33) and (374.33,184.53) .. (374.33,198.33) .. controls (374.33,212.14) and (363.14,223.33) .. (349.33,223.33) .. controls (335.53,223.33) and (324.33,212.14) .. (324.33,198.33) -- cycle ;
\draw  [color={rgb, 255:red, 0; green, 0; blue, 0 }  ,draw opacity=1 ][fill={rgb, 255:red, 155; green, 155; blue, 155 }  ,fill opacity=1 ] (323.67,62.33) .. controls (323.67,48.53) and (334.86,37.33) .. (348.67,37.33) .. controls (362.47,37.33) and (373.67,48.53) .. (373.67,62.33) .. controls (373.67,76.14) and (362.47,87.33) .. (348.67,87.33) .. controls (334.86,87.33) and (323.67,76.14) .. (323.67,62.33) -- cycle ;
\draw  [color={rgb, 255:red, 0; green, 0; blue, 0 }  ,draw opacity=1 ] (373.5,103) -- (440.5,103) -- (440.5,141.2) -- (373.5,141.2) -- cycle ;
\draw [color={rgb, 255:red, 0; green, 0; blue, 0 }  ,draw opacity=1 ][line width=0.75]    (53.8,120.6) -- (96.05,160.62) ;
\draw [shift={(97.5,162)}, rotate = 223.45] [fill={rgb, 255:red, 0; green, 0; blue, 0 }  ,fill opacity=1 ][line width=0.75]  [draw opacity=0] (8.93,-4.29) -- (0,0) -- (8.93,4.29) -- cycle    ;

\draw [color={rgb, 255:red, 0; green, 0; blue, 0 }  ,draw opacity=1 ][line width=3.75]    (53.8,120.6) -- (93.82,74.15) ;
\draw [shift={(97.73,69.6)}, rotate = 490.74] [fill={rgb, 255:red, 0; green, 0; blue, 0 }  ,fill opacity=1 ][line width=3.75]  [draw opacity=0] (22.33,-10.72) -- (0,0) -- (22.33,10.73) -- (14.83,0) -- cycle    ;

\draw [color={rgb, 255:red, 0; green, 0; blue, 0 }  ,draw opacity=1 ][line width=3.75]    (147.73,69.6) -- (193.35,106.87) ;
\draw [shift={(198,110.67)}, rotate = 219.25] [fill={rgb, 255:red, 0; green, 0; blue, 0 }  ,fill opacity=1 ][line width=3.75]  [draw opacity=0] (20.54,-9.87) -- (0,0) -- (20.54,9.87) -- cycle    ;

\draw [color={rgb, 255:red, 0; green, 0; blue, 0 }  ,draw opacity=1 ][line width=0.75]    (146.53,170) -- (196.69,112.18) ;
\draw [shift={(198,110.67)}, rotate = 490.94] [fill={rgb, 255:red, 0; green, 0; blue, 0 }  ,fill opacity=1 ][line width=0.75]  [draw opacity=0] (8.93,-4.29) -- (0,0) -- (8.93,4.29) -- cycle    ;

\draw [color={rgb, 255:red, 0; green, 0; blue, 0 }  ,draw opacity=1 ]   (247,187.67) -- (322.62,64.04) ;
\draw [shift={(323.67,62.33)}, rotate = 481.45] [fill={rgb, 255:red, 0; green, 0; blue, 0 }  ,fill opacity=1 ][line width=0.75]  [draw opacity=0] (8.93,-4.29) -- (0,0) -- (8.93,4.29) -- cycle    ;

\draw [color={rgb, 255:red, 0; green, 0; blue, 0 }  ,draw opacity=1 ]   (247,187.67) -- (322.35,198.06) ;
\draw [shift={(324.33,198.33)}, rotate = 187.85] [fill={rgb, 255:red, 0; green, 0; blue, 0 }  ,fill opacity=1 ][line width=0.75]  [draw opacity=0] (8.93,-4.29) -- (0,0) -- (8.93,4.29) -- cycle    ;

\draw [color={rgb, 255:red, 0; green, 0; blue, 0 }  ,draw opacity=1 ][line width=3.75]    (248,110.67) -- (320.39,193.81) ;
\draw [shift={(324.33,198.33)}, rotate = 228.95] [fill={rgb, 255:red, 0; green, 0; blue, 0 }  ,fill opacity=1 ][line width=3.75]  [draw opacity=0] (20.54,-9.87) -- (0,0) -- (20.54,9.87) -- cycle    ;

\draw [color={rgb, 255:red, 0; green, 0; blue, 0 }  ,draw opacity=1 ]   (247,28.67) -- (323.5,196.51) ;
\draw [shift={(324.33,198.33)}, rotate = 245.5] [fill={rgb, 255:red, 0; green, 0; blue, 0 }  ,fill opacity=1 ][line width=0.75]  [draw opacity=0] (8.93,-4.29) -- (0,0) -- (8.93,4.29) -- cycle    ;

\draw [color={rgb, 255:red, 0; green, 0; blue, 0 }  ,draw opacity=1 ]   (247,270.67) -- (322.87,199.7) ;
\draw [shift={(324.33,198.33)}, rotate = 496.91] [fill={rgb, 255:red, 0; green, 0; blue, 0 }  ,fill opacity=1 ][line width=0.75]  [draw opacity=0] (8.93,-4.29) -- (0,0) -- (8.93,4.29) -- cycle    ;

\draw [color={rgb, 255:red, 0; green, 0; blue, 0 }  ,draw opacity=1 ]   (247,270.67) -- (322.98,64.21) ;
\draw [shift={(323.67,62.33)}, rotate = 470.2] [fill={rgb, 255:red, 0; green, 0; blue, 0 }  ,fill opacity=1 ][line width=0.75]  [draw opacity=0] (8.93,-4.29) -- (0,0) -- (8.93,4.29) -- cycle    ;

\draw [color={rgb, 255:red, 0; green, 0; blue, 0 }  ,draw opacity=1 ]   (247,28.67) -- (321.84,61.53) ;
\draw [shift={(323.67,62.33)}, rotate = 203.71] [fill={rgb, 255:red, 0; green, 0; blue, 0 }  ,fill opacity=1 ][line width=0.75]  [draw opacity=0] (8.93,-4.29) -- (0,0) -- (8.93,4.29) -- cycle    ;

\draw [color={rgb, 255:red, 0; green, 0; blue, 0 }  ,draw opacity=1 ][line width=4.5]    (248,110.67) -- (317.77,66.1) ;
\draw [shift={(323.67,62.33)}, rotate = 507.43] [fill={rgb, 255:red, 0; green, 0; blue, 0 }  ,fill opacity=1 ][line width=4.5]  [draw opacity=0] (24.11,-11.58) -- (0,0) -- (24.11,11.58) -- cycle    ;

\draw [color={rgb, 255:red, 0; green, 0; blue, 0 }  ,draw opacity=1 ] [dash pattern={on 0.84pt off 2.51pt}]  (374.33,198.33) -- (411.5,145) ;

\draw [color={rgb, 255:red, 0; green, 0; blue, 0 }  ,draw opacity=1 ] [dash pattern={on 0.84pt off 2.51pt}]  (373.67,62.33) -- (411.5,103) ;

\draw (407,122.1) node [scale=1,color={rgb, 255:red, 0; green, 0; blue, 0 }  ,opacity=1 ] [align=left] {Rewards};
\draw (567,153) node [scale=1.44,color={rgb, 255:red, 0; green, 0; blue, 0 }  ,opacity=1 ] [align=left] {\textbf{State variables:}\\W - Worker number\\S - Supply depot number\\B - barracks \ number\\E - enemay location\\$A_n$ - Ally unit number\\$A_h$ - Ally unit health\\$A_l$ - Ally unit location\\$D_u$ - Destoryed units\\$D_b$ - Destroyed buildings\\\textbf{Actions:}\\$A_s$ - build supply depot\\$A_b$ - build barracks\\$A_m$ - train offensive unit\\$A_a$ - attack};
\draw (62,76) node [scale=1.7280000000000002,color={rgb, 255:red, 0; green, 0; blue, 0 }  ,opacity=1 ]  {$A_{s}$};
\draw (28.8,120.6) node [scale=1.7280000000000002,color={rgb, 255:red, 0; green, 0; blue, 0 }  ,opacity=1 ]  {$W$};
\draw (122.73,69.6) node [scale=1.7280000000000002,color={rgb, 255:red, 0; green, 0; blue, 0 }  ,opacity=1 ]  {$S$};
\draw (121.53,170) node [scale=1.7280000000000002,color={rgb, 255:red, 0; green, 0; blue, 0 }  ,opacity=1 ]  {$B$};
\draw (223,110.67) node [scale=1.7280000000000002,color={rgb, 255:red, 0; green, 0; blue, 0 }  ,opacity=1 ]  {$A_{n}$};
\draw (222,28.67) node [scale=1.7280000000000002,color={rgb, 255:red, 0; green, 0; blue, 0 }  ,opacity=1 ]  {$E_{l}$};
\draw (222,187.67) node [scale=1.7280000000000002,color={rgb, 255:red, 0; green, 0; blue, 0 }  ,opacity=1 ]  {$A_{h}$};
\draw (222,270.67) node [scale=1.7280000000000002,color={rgb, 255:red, 0; green, 0; blue, 0 }  ,opacity=1 ]  {$A_{l}$};
\draw (349.33,198.33) node [scale=1.44,color={rgb, 255:red, 0; green, 0; blue, 0 }  ,opacity=1 ]  {$D_{b}$};
\draw (348.67,62.33) node [scale=1.44,color={rgb, 255:red, 0; green, 0; blue, 0 }  ,opacity=1 ]  {$D_{u}$};
\draw (68,153) node [scale=1.7280000000000002,color={rgb, 255:red, 0; green, 0; blue, 0 }  ,opacity=1 ]  {$A_{b}$};
\draw (174,67) node [scale=1.7280000000000002,color={rgb, 255:red, 0; green, 0; blue, 0 }  ,opacity=1 ]  {$A_{m}$};
\draw (172,161) node [scale=1.7280000000000002,color={rgb, 255:red, 0; green, 0; blue, 0 }  ,opacity=1 ]  {$A_{m}$};
\draw (270,20) node [scale=1.7280000000000002,color={rgb, 255:red, 0; green, 0; blue, 0 }  ,opacity=1 ]  {$A_{a}$};
\draw (265,50) node [scale=1.7280000000000002,color={rgb, 255:red, 0; green, 0; blue, 0 }  ,opacity=1 ]  {$A_{a}$};
\draw (263,82) node [scale=1.7280000000000002,color={rgb, 255:red, 0; green, 0; blue, 0 }  ,opacity=1 ]  {$A_{a}$};
\draw (265,128) node [scale=1.7280000000000002,color={rgb, 255:red, 0; green, 0; blue, 0 }  ,opacity=1 ]  {$A_{a}$};
\draw (263,153) node [scale=1.7280000000000002,color={rgb, 255:red, 0; green, 0; blue, 0 }  ,opacity=1 ]  {$A_{a}$};
\draw (265,189) node [scale=1.7280000000000002,color={rgb, 255:red, 0; green, 0; blue, 0 }  ,opacity=1 ]  {$A_{a}$};
\draw (273,272) node [scale=1.7280000000000002,color={rgb, 255:red, 0; green, 0; blue, 0 }  ,opacity=1 ]  {$A_{a}$};
\draw (264,224) node [scale=1.7280000000000002,color={rgb, 255:red, 0; green, 0; blue, 0 }  ,opacity=1 ]  {$A_{a}$};

\end{tikzpicture}

}
\caption{Action influence graph of a Starcraft II agent} \label{fig1}
\end{figure}
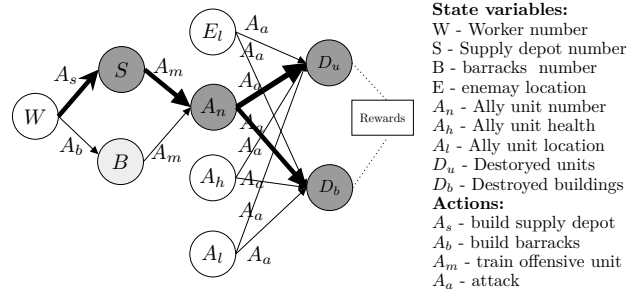

Informally, this defines a complete explanan for action $a$ as the complete causal chain from action $a$ to any future reward that it can receive. From Figure \ref{fig1}, the causal chain for action $\mathcalc{A_s}$ is depicted in bold edges, and the extracted explanan tuple $ \left ( \left [S=s \right ], \left [ A_n=a_n\right ], \left [  D_u=d_u, D_b=d_b\right ] \right )$ is shown as darkened nodes. We use depth-first search to traverse the graph until all the sink nodes are reached from the head node of the action edge.







\subsection{`Why?' Questions}

~\citeauthor{lim2009and}~\shortcite{lim2009and} found that the most demanded explanatory questions are \emph{Why} and \emph{Why not} questions. To this end, we focus on explanation generation for \emph{why} and \emph{why not} questions in this paper.


\subsubsection{Minimally Complete Explanations}

Striking a balance between \emph{complete} and \emph{minimal} explanations depend on the epistemic state of the explainee~\cite{miller2018explanation}. In this paper, we assume that we know nothing about the epistemic state of the explainee. 

Recall from the definition of \emph{explanans} (Definition \ref{def5}), a `complete' explanation would include \emph{explanans} of all the intermediate nodes between the head and reward node of the causal chain. Clearly, for a large graph, this risks overwhelming the explainee. For this reason, we define \emph{minimally complete} explanations.

\citeauthor{mcclure1997you}~\shortcite{mcclure1997you} show that referring to the goal as being the most important for explaining actions. In our causal models, the rewards are the `goals', but these alone do not form meaningful explanations because they are merely numbers. We define the human interpretable `goal' using the variables in the predecessor nodes of the rewards. These define the immediate causes of the reward, and therefore which states will result in rewards. However, this alone is only a longer-term motivation for taking an action. As such, we also include the head node of the action edge as the immediate reason for doing the action. We use this model to define our \emph{minimally complete} explanations. 

\begin{definition}\label{def7}
A \emph{minimally complete} explanation is a tuple $\left (  \vec{X_r}=\vec{x_r}, \vec{X_h}=\vec{x_h}, \vec{X_p}=\vec{x_p}\right )$, in which $ \vec{X_r}=\vec{x_r}$ and $\vec{X_h}=\vec{x_h}$ do not change from Definition \ref{def5}, and $\vec{X_p}=\vec{x_p}$ is the vector of variables that are immediate predecessors of any variable in ${X_r}$ within the causal chain, with $\vec{x_p}$ the values in the actual instantiation.    
\end{definition}

Informally,  for a complete causal chain, we take the first and last arcs of the causal chain, with their source and destination nodes, omitting intermediate nodes, as the minimal explanation. From Figure \ref{fig1}, for the action $A_s$, the minimally complete explanation is just the complete explanation, as there are no intermediate nodes.

Clearly, one could define other heuristics to decide which intermediate nodes to use as explanations, such as the knowledge of the explainee. However, for the purposes of this paper, we use this simple definition.

\subsection{`Why not?' Questions}

\emph{Why not} questions let the explainee ask why an event has not occurred, thus allowing \emph{counterfactuals} to be explained; something that is known to be a powerful explanation mechanism \cite{miller2018explanation,byrne2019counterfactuals}. Our model generates counterfactual explanations by comparing causal chains of the actual event occurred and the \emph{explanandum} (counterfactual action). First, we define a \emph{counterfactual instantiation} that specifies the optimal state variable values under which the counterfactual action $B$ would be chosen.


\begin{definition}
\label{counterfactual}
A \emph{counterfactual instantiation} for a counterfactual action $B$ is a model $M_{\vec{Z}\leftarrow \vec{S_Z}}$, where $\vec{Z}$ gives the instantiation of all predecessor variables of action $B$ with current state values \emph{and} the instantiation of all successor nodes (of $B$) of the causal chain by forward simulating, using the structural equations.
\end{definition}

Informally, this gives the `optimal' conditions (according to the action influence model) under which we would select counterfactual action $B$, simulated through structural equations. We unravel this further in the Example \ref{ex1} discussion using the Starcraft II scenario.


In the following definition, we use $\vec{X}=\vec{x}$ to represent the tuple $\left (\vec{X_p}=\vec{x_p}, \vec{X_h}=\vec{x_h}, \vec{X_r}=\vec{x_r} \right )$, and similar for $\vec{Y}=\vec{y}$ for readability.

\begin{definition}
\label{def9}
Given a minimally complete explanation $\vec{X}=\vec{x}$ for action $A$ under the actual instantation, and a minimally complete explanation $\vec{Y}=\vec{y}$ for action $B$ under the counterfactual instantiation $M_{\vec{Z}\leftarrow \vec{S_Z}}$ (from Definition \ref{counterfactual}),
we define a \emph{minimally complete contrastive explanation} as the tuple $(\vec{X'}=\vec{x'}, \vec{Y'}=\vec{y'}, \vec{X_r}=\vec{x_r})$ such that $\vec{X'}$ is the maximal set of variables in $\vec{X}$ in which $(\vec{X'}=\vec{x'}) \cap (\vec{Y'}=\vec{y'}) \neq \emptyset$, where $\vec{x'}$ is then contrasted with $\vec{y'}$. That is, we only explain things that are different between the actual and counterfactual. This corresponds to the  \emph{difference condition} \cite{miller2018contrastive}. And $\vec{X_r}$ gives the reward nodes of action $A$.
\end{definition}

Intuitively, a contrastive explanation extracts the actual causal chain for the taken action $A$, and the counterfactual causal chain for the $B$, and finds the differences.



\begin{example}
\label{ex1} 
Consider the question, asking why a Starcraft II agent built supply depots, rather than choosing to build barracks:

\begin{center}
\noindent
\begin{tabular}{@{}l@{~~}p{0.35\textwidth}@{}}
\emph{Question} & Why not $build\_barracks$ ($A_b$)?\\
\emph{Explanation} & Because it is more desirable to do action build\_supply\_depot ($A_s$) to have more Supply Depots ($S$) as the goal is to have more Destroyed Units ($D_u$) and Destroyed buildings ($D_b$).
\end{tabular}
\end{center}
\end{example}

First we get the \emph{actual instantiation} $m = \left [  W=12, S=1, B=2, A_n=22, D_u=10, D_b=7\right ]$ (instantiation should include all variables in the current state, only the required ones are shown for readability). The causal chain for the \emph{actual} action `why $A_s$?' would be as in Figure 1, and for the \emph{counterfactual} action `why not $A_b$?', the causal chain nodes would be $B \rightarrow A_n \rightarrow [D_u, D_b]$. We then get the \emph{counterfactual instantiation} $m' = \left [  W=12, S=3, B=2, A_n=22, D_u=10, D_b=7\right ]$ using Definition \ref{counterfactual}. Applying the difference condition here, we obtain the minimally complete contrastive explanation (from Definition \ref{def9}) as the tuple $ \left ( \left [S=1\right ],\left [S=3\right ], [D_u=10, D_b=7] \right )$ and  contrast $\left [S=1\right ]$ with  $\left [S=3\right ]$ to obtain the explanation of Example \ref{ex1} (generated using a simple NLP template).

\subsection{Learning Structural Causal Equations}


Our approach so far relies on knowing the structural model, in particular, to determine the effects of counterfactual actions. \emph{Why not} questions are inherently counterfactual \cite{balke1995counterfactuals}, and having just the policy of an agent is not enough to generate explanations as counterfactuals refers to \emph{possible} worlds that did not happen. Consider the Example \ref{ex1}, to generate this explanation, the optimal/maximum value of the state variable $S$ is needed in the given time instance.  



However, in model-free reinforcement learning, such environment dynamics are not known. And learning a model of the environment is a difficult problem. Though, when given a graph of causal relations between variables, learning a set of structural equations that are approximate yet `good enough' to give counterfactual explanations maybe feasible. 


To this end, we assume that a DAG specifying causal direction between variables is given, and learn the structural equations as multivariate regression models during the training phase of the RL agent. We perform experience replay~\cite{mnih2015human} by saving $e_t = (s_t, a_t, r_t, s_{t+1})$ at each time step $t$ in a data set $D_t = \left \{ e_1,..., e_t \right \}$. Then we update the sub-set of structural equations $F_{X.A}$ using a regression learner $\mathbb{\widehat{L}}_{(s, a, r, s') \sim U(D) }$, in that we \emph{only} update structural equations associated with the specific action in the experience frame, drawn uniformly as mini-batches from $D$. For example, from Figure \ref{fig1}, for any experience frame with the action $A_s$, only the equation $\mathcalc{F_{S.A_s}(W)}$ will be updated. Any regression learner can be used as the learning model $\mathbb{\widehat{L}}$, such as multi-layer perceptron regressors. 


While this approach may seem similar to learning environment dynamics of model-based RL methods, we only learn the structural equations, and we are only after an approximation that is good enough for explaining instances. Thus they can be approximate but still useful for explanation. Further, specifying the assumptions about the causal direction  between variables is a much easier problem to encode by hand, and can be tested with the data. 








\section{Computational Evaluation}
\label{sec:computational}

We evaluate \emph{action influence models} in 5 OpenAI RL benchmark domains~\cite{openai} and in the Starcraft II domain. The goal of this evaluation is to determine if learning action influence models leads to models that are faithful to the problem. Task prediction accuracy and training time for the structural causal equations are measured. The purpose of task prediction is to evaluate if the model is accurate enough to predict what an agent will do next, under the assumption that if it is not, then the model will not be of use to a human explainee. 

We computationally simulate task prediction using Algorithm~\ref{alg:algorithm2}. Here we instantiate all the equations (which are the set of regression models $\mathcal{L}$) with the values of the current state $S$ of the agent. We identify the equation that has maximum difference with the predicted state variable value and the actual, then get the action associated with it. This is informed by the reasoning that the agent will try to follow the optimal policy, and the action with the biggest impact to correct the policy will be executed. The impact is measured by the above mentioned difference.  This is itself an approximation, but is a useful guide for task prediction.


\begin{algorithm}[tb]
\small
\caption{Task Prediction:Action Influence Model}
\label{alg:algorithm2}
\textbf{Input}: trained regression models $\mathcal{L}$, current state $S_t$\\
\textbf{Output}: predicted action $a$
\begin{algorithmic}[1] 
\State $\vec{F_p} \leftarrow \left [  \right ]$ ; vector of predicted difference

\For{every $\widehat{L} \in \mathcal{L}$ }
\State ${P_y} \leftarrow \widehat{L} \cdot predict (S_{x.t}) $; predict variable $S_y$ at $S_{t+1}$

\State $\vec{F_p} \leftarrow | S_y - P_y|$; difference with actual $S_y$ value

\EndFor

\State \Return{$max\left ( \vec{F_p} \right )\cdot getAction()$}

\end{algorithmic}
\end{algorithm}

We use linear SGD regression (LR), decision tree regression (DT) and multilayer perceptron regression (MLP) as the learners that approximate the structural equations. We choose benchmark domains based on varying levels of complexity, size (state features/number of actions) and train them using various RL algorithms to demonstrate the robustness of the model. Table~\ref{tablecompute} summarises the results of task prediction and time taken to train the structural equations given the replay data.

Overall, the results show the model did a reasonable job of task prediction, providing evidence that this could be useful for  explanations.
Domains that have a clear causal structure (e.g Starcraft) performs best in task prediction. Considering the performance cost it incurs, there was little  gained by using MLP to approximate the equations, where in most cases linear regression is adequate. Apart from the BipedalWalker domain, our model performs well in task prediction with a negligible performance hit. The bipedalWalker domain has continuous actions, which our current model cannot handle accurately. We plan to extend our model to continuous actions in future work.

\begin{table}

\resizebox{\columnwidth}{!}{%
  \begin{tabular}{p{3.2cm}@{ }r r r @{ }r r @{ }r @{ }r r r r}
    \toprule
    \multirow{2}{*}{Env - RL} &
      \multicolumn{1}{c}{} &
      \multicolumn{3}{c}{Accuracy (\%)} &
      \multicolumn{3}{c}{Performance (s)} \\
      \cmidrule(lr){3-5}\cmidrule(l){6-8}
       & {Size} & {LR} & {DT} & {MLP} & {LR} & {DT} & {MLP} \\
      \midrule
    Cartpole-PG  & 4/2 &     83.8 & 81.6 & 86.0      & 0.007 & 0.018 & 0.03 \\
    MountainCar-DQN & 3/3 &    69.7 & 57.8 & 69.6     & 0.020 & 0.037 & 0.32  \\
    Taxi-SARSA &  4/6 &       68.2 & 74.2 & 67.9       & 0.001 & 0.001 & 0.49  \\
    LunarLander-DDQN  & 8/4     & 68.4 & 63.7 & 72.1      & 0.002 & 0.002 & 0.33  \\
    BipedalWalker-PPO  & 14/4      & 56.9 & 56.4 & 56.7      & 0.010 & 0.015 & 0.41   \\
    Starcraft-A2C  & 9/4       & 94.7 & 91.8 & 91.4       & 0.144 & 0.025 & 3.33 \\
    \bottomrule
  \end{tabular}
  }
  \caption{Action influence model evaluation in 6 benchmark reinforcement learning domains (using different RL algorithms, PG, DQN etc.), measuring mean task prediction accuracy and training time of the structural causal equations in 100 episodes after training.}
  \label{tablecompute}
\end{table}



\section{Empirical Evaluation: Human Study}
\label{sec:experiment}

A human-grounded evaluation is essential to evaluate the explainability of a system, thus we carry out human-subject experiments involving explaining RL agents. We present two main hypotheses for the empirical evaluation; \textbf{H1)} Causal-model-based explanations build better mental models of the agent leading to a better \textbf{understanding} of its strategies (We make the assumption here that there is no intermediate effect on the mental model from other sources); and \textbf{H2)} Better understanding of an agent's strategies promotes \textbf{trust} in the agent. 


\subsubsection{Methodology:}

We use StarCraft II, a real-time strategy game and a popular RL environment~\cite{vinyals2017starcraft} as the domain. We implemented a RL agent for our experiment that competes in the default map.



To evaluate  hypothesis (H1), we use the method of \emph{task prediction}~\cite{hoffman2018metrics}. Task prediction can provide a quick view of the explainee's mental model formed through explanations, where the task is for the participant to predict `What will the agent do next?'. We use the 5-point Likert \emph{Explanation Satisfaction Scale} developed by~\citeauthor{hoffman2018metrics}~\shortcite[p.39]{hoffman2018metrics} to measure the subjective quality of explanations. To evaluate  hypothesis (H2), we use the 5-point Likert \emph{Trust Scale} of ~\citeauthor{hoffman2018metrics}~\shortcite[p.49]{hoffman2018metrics}. We obtained ethics approval from The University of Melbourne human research ethics committee (ID-1953619.1).

\subsubsection{Experiment Design:}
We use a recording of a full gameplay video (22 min) with the RL agents playing against in-game bot AI. The experiment has 4 phases.

Phase 1 involves collecting demographic information and training the participants. 
Using five gameplay video clips, the participant is trained to understand and differentiate the actions of the agent.

In phase 2, a clip of the gameplay video (15 sec) is played in a web-based UI, with a textual description of the scene. The participant can  select the question type (why/why not) and the action, which together forms a question `Why/Why not \emph{action A}?'. Then, the textual explanation  for the question  with a figure of the relevant sub-graph of the agent's action influence graph is displayed. Explanations are pre-generated from our implemented algorithm. The participant can ask \textbf{multiple} questions in a single gameplay video. After every gameplay video, the participant completes the \emph{Explanation Satisfaction Scale}. This process is repeated so we have data for each participant from five videos. 

The third phase measures the \emph{understanding} the explainee has after seeing the gameplay and the explanations. We measure understanding using the task prediction method as follows: the participant is presented with another gameplay video (10 sec), and presented with three selections of textual descriptions of what \emph{action} the agent will do in \emph{next} step; the participant selects an option, which includes `I don't know'. We expect the participant is projecting forward the \emph{local strategy} of the agent using their mental model. This mental model is formed through (or helped by)  explanations seen in phase 1. This process is repeated for 8 tasks. In 4 of the task predictions, the behaviour is explainable using a causal chain previously seen in the training, but with different variable values. In the other 4 tasks, the behaviour is novel, but can be inferred by combining causal chains from different training tasks. In the fourth phase, the participant completes a 5-point \emph{Trust Scale}.




We conducted the experiments on \emph{Amazon MTurk}, a crowd-sourcing platform popular for obtaining data for human-subject experiments~\cite{buhrmester2011amazon}. The experiment was fully implemented in an interactive web-based environment. We excluded noisy data of users in 3 ways. First, we tested participants to ensure they had learnt about the agent's actions by prompting them to identify them. If the participant failed this, the experiment did not proceed. Then, for participants who completed, we omitted their data from analysis based on two criteria: 1) if the threshold of the time the participant spent on viewing explanations and answering tasks is below a few seconds, which was deemed too short to learn anything useful; and 2) if the participant's textual responses to explain their task prediction choice were gibberish text or a 1-2 word response, as this indicated lack of engagement and care in the task. We controlled for language by only recruiting participants from the US.

\subsubsection{Experiment Parameters:} The experiment was run with 4 independent variables. We tested abstract (C) and detailed (D) versions of our action influence models and  2 baseline models described below: 1) Gameplay video without any  explanations (N); 2) Relevant variable explanations (R). These explanations are generated using state  relevant variables using \emph{template 1} of~\citeauthor{khan2009minimal} \shortcite[p.3]{khan2009minimal} and visualized through a state-action graph, e.g `Action \emph{A} is likely to increase \emph{relevant variable P}'; 3) Detailed action influence model explanations, where the causal graph is augmented to include atomic actions. 



We ran experiments for \textbf{120} participants, allocated evenly to the independent variables. Each experiment ran for approximately 40 minutes. We scored each participant on task prediction, 2 points for a correct prediction; 1 for responding `I don't know' and 0 for an incorrect prediction for a total of 16 points. Scores were tallied. We compensated each participant with 8.5USD. Of the 120 participants, 36 were female, 82 male and 2 were not given. Participants were aged between 19 to 59 ($\mu=34.2$) and had an average self-rated gaming experience and Starcraft II experience of 3.38 and 2.02 (5-point Likert) respectively.




\begin{figure*}[!t]
\minipage{0.32\textwidth}
  \includegraphics[width=\linewidth]{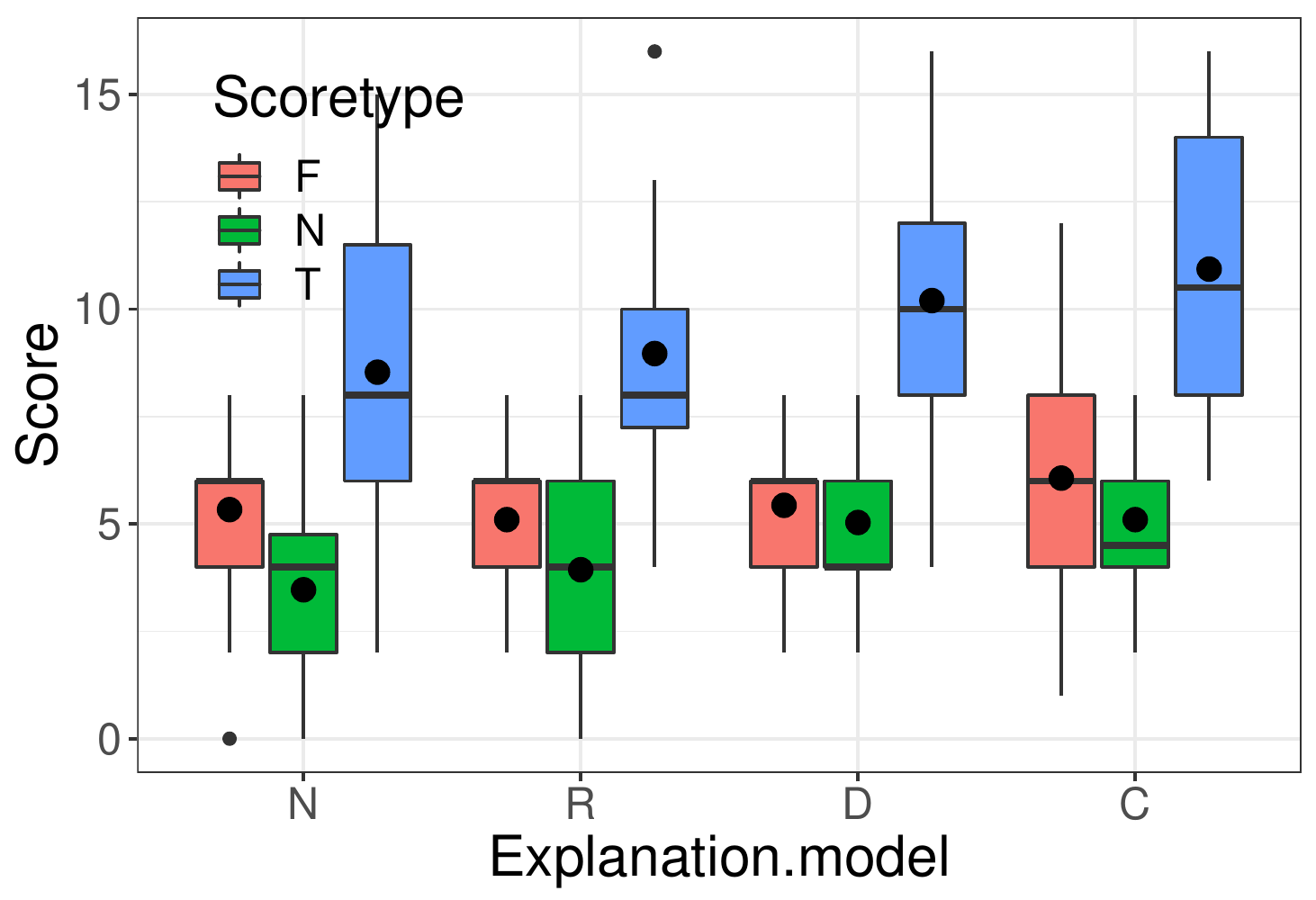}
  \caption{Box plot of task prediction scores of explanation models, T=total score, F=familiar score, N=novel score (higher is better, means represented as bold dots).}\label{fig2}
\endminipage\hfill
\minipage{0.32\textwidth}
  \includegraphics[width=\linewidth]{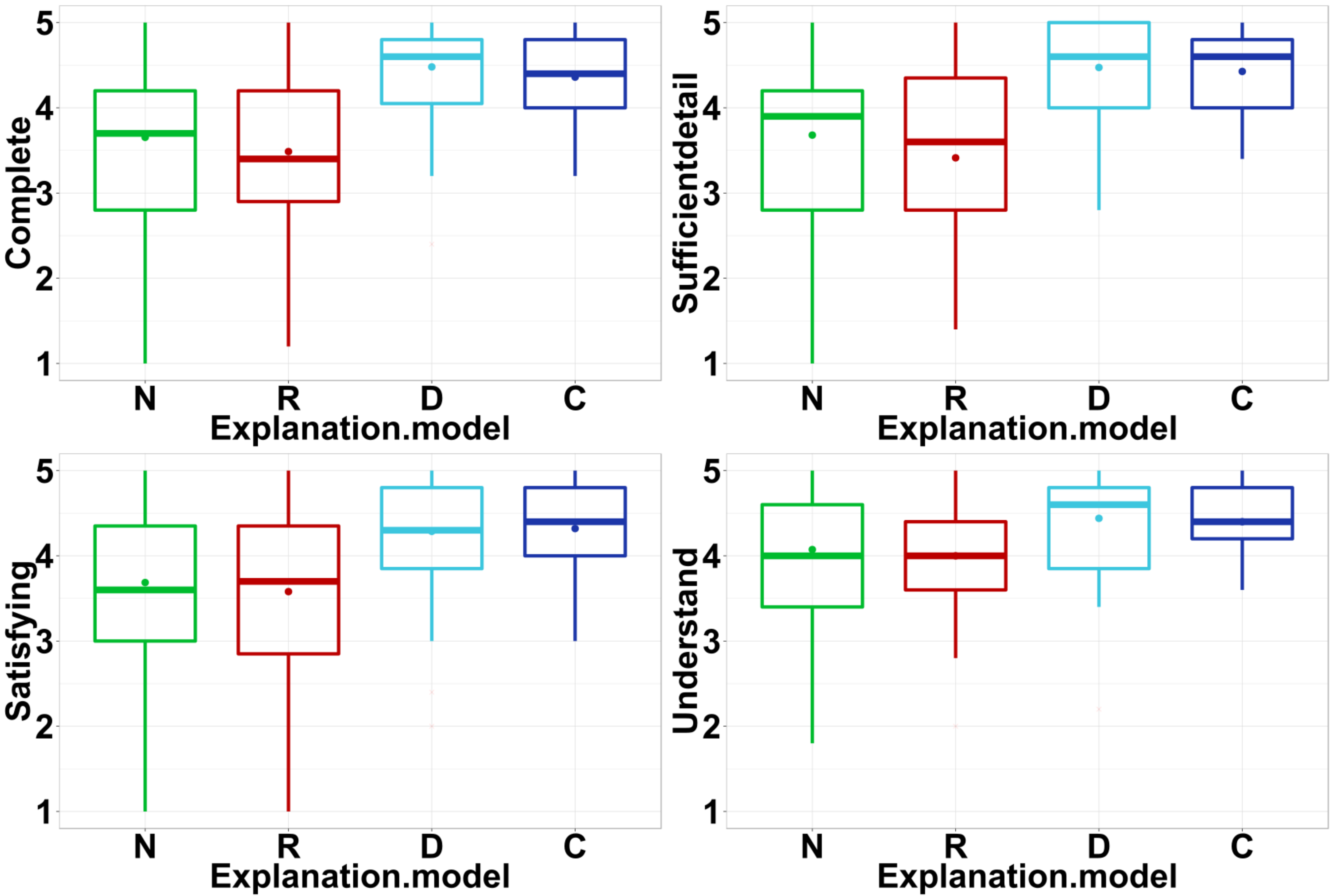}
  \caption{Box plot of explanation quality (likert scale 1-5, higher is better, means represented as dots).}\label{fig3}
\endminipage\hfill
\minipage{0.32\textwidth}%
  \includegraphics[width=\linewidth]{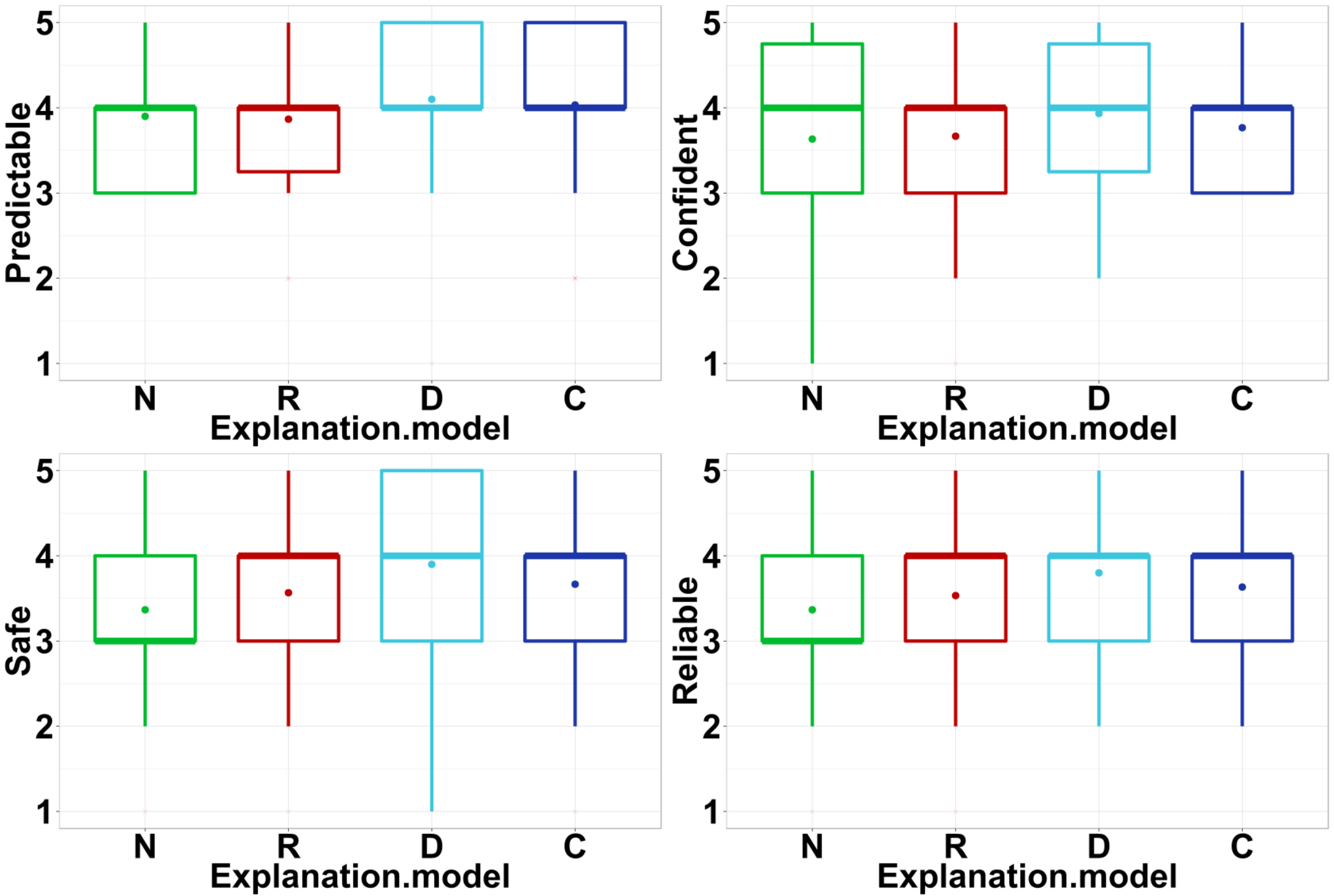}
  \caption{Box plot of trust (likert scale 1-5, higher is better, means represented as dots).}\label{fig4}
\endminipage
\end{figure*}

\subsection{Results}

\subsubsection{Task Prediction:}
For the first hypothesis, the corresponding null and alternative hypotheses are: 1) $H_0 : \mu_C = \mu_R = \mu_D = \mu_N$; 2) $H_1 : \mu_C \geq   \mu_R$; 3) $H_2 : \mu_C \geq  \mu_D$; 4) $H_3 : \mu_C \geq \mu_N$, where abstract causal explanations (our model), detailed causal explanations, relevant variable explanations, and no explanations are given by C, D, R, and N respectively. 



We conduct one-way ANOVA (Figure \ref{fig2} illustrates the task score variance with explanation models). We obtained a p-value of \textbf{0.003} ($\mu_C=10.90$, $\mu_D=10.20$, $\mu_R=8.97$, $\mu_N=8.53$), thus we conclude there are significant differences between models on task prediction scores. We performed Tukey multiple pairwise-comparisons to obtain the significance between groups. From Table \ref{table1}, the differences between the causal explanation model paired with other explanation models are significant for C-R and C-N pairs with p-values of \textbf{0.006} and \textbf{0.034}. Additionally, we calculate the effect of the number of questions on the score, and obtain no statistical correlation using a correlation test (number of questions vs score, p = 0.33, model C) among same models. Because participants could select ``I don't know'' and receive 8 out of 16, we also  further analyse scores based on 2 = `correct', 0 = `incorrect or `I don't know', and obtain results that are still significant (p=0.004), means (C=10.90, D=10.10, R=8.93, N=8.47), for model pairs (C-N p=0.005, C-R p=0.035). We conducted Pearson's Chi-Square as a non-parametric test on the task prediction scores, which showed significant results (p-value = 0.008, X-squared = 17.281).

Therefore we reject $H_0$ and $H_2$ and accept all other alternative hypotheses. Our results show that causal model explanations lead to a significantly better \textbf{understanding} of agent's strategies than the 2 baselines we evaluated, especially against previous models of relevant explanations. Participants did slightly worse on tasks with novel behaviour.
 
\begin{table}[!t]
\centering
\small
\begin{tabular}{lrrrr}  
\toprule
Model pair  & mean-diff & lwr & upr & p-value \\
\midrule
C - N   & \textbf{2.400}  & 0.534  &   4.265 & \textbf{0.006}\\
C - R   & \textbf{1.966}  & 0.101  &   3.832 & \textbf{0.034}\\
D - N   & 1.666  & -0.198  &  3.532  & 0.097\\
D - R   & 1.233  & -0.632  &   3.098 &  0.316\\
C - D   & \textbf{0.733}  & -1.132  &   2.598  &  \textbf{0.735}\\
R - N   & 0.433  & -1.432 &   2.298  &  0.930\\
\bottomrule
\end{tabular}
\caption{Pairwise-comparisons of explanation models of task prediction scores (higher positive diff is better)}
\label{table1}
\end{table}

\begin{table}[!t]
\centering
\small
\begin{tabular}{lrrrr}

\toprule

Metric & Mdl-pair & Mean-dif & Median-dif & p-val \\\midrule

\multirow{2}{*}{Complete} & C-N  & 0.707  & 0.700 & 0.061\\
   & C-R  & 0.873 & 1.000 & \textbf{0.012} \\
\hline
  
\multirow{2}{*}{Sufficient} & C-N  & 0.746  & 0.700 & \textbf{0.039}\\
   & C-R & 1.013  & 1.000 & \textbf{0.002 }\\\hline
  
\multirow{2}{*}{Satisfying} & C-N & 0.633 & 0.800 & 0.082\\
   & C-R & 0.740 & 0.700 & \textbf{0.029} \\
\hline
\multirow{2}{*}{Understand} & C-N & 0.326 & 0.400  & 0.497\\
   & C-R & 0.400 & 0.400 & 0.316 \\
\bottomrule

\end{tabular}
\caption{Explanation quality (likert scale data 1-5)}
\label{table2}
\end{table}


\subsubsection{Explanation Quality:}
Figure \ref{fig3} depicts the likert scale data of explanation metrics (understand, satisfying, sufficient detail and complete) for aggregated  video explanations of explanation models. As before we performed a pair-wise ANOVA test, results are summarised in Table \ref{table2}. Our model obtained statistically significant results and outperformed the benchmark `relevant explanation' (R) for all metrics except `Understand'.



\subsubsection{Trust:}
For the second main hypothesis (H2) that investigate whether explanation models promote trust, the obtained p-values for trust metrics \emph{confident}, \emph{predictable}, \emph{reliable} and \emph{safe} were not statistically significant (using pair-wise ANOVA). Although the difference is not significant we can see causal models have high means and medians (see Figure \ref{fig4}). We conclude that while the explanation quality and scores are significantly better for our model, to promote trust further interaction is necessary; or perhaps our RL agent is simply not a trustworthy Starcraft II player.

We further analysed self-reported demographic data to see if there is a correlation between task prediction scores and self-reported Starcraft II experience level (5-point Likert). Pearson's correlation test was not significant (p=0.45) thus we conclude there is no correlation between scores and experience level. This can possibly be attributed to our Starcraft II scenario differing from the standard game.  

A limitation of our experiment is that we made a strong linearity assumption for Starcraft II, which enabled linear regression to learn SCMs for a relatively small number (9) of state variables.





\section{Conclusion}

In this paper, we introduced \emph{action influence} models for model-free reinforcement learning agents. Our approach learns a structural causal model (SCM) during reinforcement learning and has the ability to generate explanations for \emph{why} and \emph{why not} questions by counterfactual analysis of the learned SCM. We computationally evaluated our model in 6 benchmark RL domains on \emph{task prediction}. We then conducted a human study (\textbf{n=120}) to evaluate our model on 1) task prediction, 2) explanation `goodness' and 3) trust. Results show that our model performs significantly better in the first 2 evaluation criteria. One weakness of our approach is that the causal model must be given beforehand. Future work includes using epistemic knowledge of the explainee to provide  explanations that are more targeted, and extending the model to continuous domains.







\section{Acknowledgment}

This research  was supported by the University of Melbourne research scholarship (MRS) and by Australian Research Council Discovery Grant DP190103414: \textit{Explanation in Artificial Intelligence: A Human-Centred Approach}. The authors also thank Fatma Faruq for the valuable feedback on early drafts of this paper.

\bibliographystyle{aaai}
\bibliography{main}

\end{document}